\documentclass[11pt, a4paper, logo, onecolumn, copyright]{googledeepmind}

\usepackage[authoryear, sort&compress, round]{natbib}
\usepackage{array}
\usepackage{multirow}
\usepackage{caption}
\usepackage{tikz}
\usetikzlibrary{arrows.meta,positioning,shapes.geometric,calc}

\usetikzlibrary{arrows,positioning}
\usepackage[titles]{tocloft}
\usepackage{amsthm}
\usepackage{blindtext}
\usepackage{mathtools}

\theoremstyle{definition}
\newtheorem{defn}{Definition}[]

\theoremstyle{conjecture}
\newtheorem{conj}{Conjecture}[]

\DeclareMathOperator*{\argmax}{argmax}

\usepackage{tcolorbox}
\usepackage{wrapfig}

\title{A Theory of Appropriateness That Accounts for Norms of Rationality}

\reportnumber{} 

\author[1]{Joel Z.~Leibo}
\author[1]{Alexander Sasha Vezhnevets}
\author[3]{Manfred Diaz}
\author[1]{John P.~Agapiou}
\author[1, 2]{William A. Cunningham}
\author[1]{Peter Sunehag}
\author[1]{Logan Cross}
\author[1]{Raphael Koster}
\author[1]{Stanley M.~Bileschi}
\author[1]{Minsuk Chang}
\author[4]{Iyad Rahwan}
\author[1]{Simon Osindero}
\author[5, 6, 7]{James A.~Evans}

\affil[1]{Google DeepMind}
\affil[2]{University of Toronto}
\affil[3]{Mila - Qu\'{e}bec AI Institute}
\affil[4]{Max Planck Institute for Human Development}
\affil[5]{Google, Paradigms of Intelligence team}
\affil[6]{University of Chicago}
\affil[7]{Santa Fe Institute}

\begin{abstract}
We propose a society-first theory of normative appropriateness where individuals, modeled as pre-trained actors with cognitive architectures analogous to Large Language Models (LLMs), generate behavior via predictive pattern completion. Our theory posits that individuals act by completing distributed symbolic patterns based on context, answering questions such as ``What does a person such as I do in a situation such as this?''. This sense-making mechanism provides a parsimonious account of the key features of human norms---their context-dependence, arbitrariness, automaticity, dynamism, and their support from social sanctioning. It challenges rational-choice theories of social norms by accounting for their key features without needing to exogenously posit scalar rewards or preference relations. By distinguishing between explicit norms, which we associate with in-context adaptation, and implicit norms, which we associate with long-term memory, the theory reconceptualizes several foundational ideas in cognitive science. In particular, it gives an alternative account to the data traditionally seen as supporting dual-process models, and it flips the role of rationality, allowing us to construe it as adherence to culturally-contingent justification standards.
\end{abstract}

\begin{document}

\maketitle

{
\tableofcontents
}



\section{Introduction}

A foundational question for the sciences of the mind and society is how social order emerges. Within sociology, this question has long been framed by a core dichotomy between \textbf{`individual-first'} and \textbf{`society-first'} (i.e.~structural) explanations---a divide as central to sociology as the `nature versus nurture', `symbolic versus distributed', or `intuitive vs deliberative' dichotomies are to cognitive science. The individual-first perspective, often termed \textbf{methodological individualism}, seeks to explain social phenomena by starting with the individual. It typically models under-socialized agents who learn either mostly or entirely from direct interaction with their environment. Society is then derived as the aggregate consequence of their individual actions. Cognitive science, in its effort to build from the ground up, has overwhelmingly adopted this explanatory pattern.

This paper, in contrast, develops a society-first theory of behavior and social order. In our framework, society is presupposed in the decision logic of the individual. Instead of assuming blank-slate learners, our theory models individual minds as Large Language Models (LLMs). Crucially, these LLMs are pre-trained on human language and cultural products. This pre-training means our agents do not begin as naive individuals from which society must be derived. Rather, they are initialized with substantial, historically-contingent cultural knowledge from the outset. Social phenomena are then generated by the interactions of these already culturally adept agents, enabling co-constitution: individuals are animated by society, and through their interactions, they collectively enact and constitute society. This co-constitution is perhaps the core meta-theoretical framework of sociology, and has been variously theorized in terms of macro- and micro-social structures ~\cite{giddens1984elements, coleman1990foundations}, multi-scale cultures like Bourdieu's habitus ~\cite{bourdieu2020outline}, and the intersection of biography and history ~\cite{mills2000sociological}. While this approach has less to say about the primordial origins of culture or the cognitive development of individuals prior to acquiring it, the gain in expressivity for simulating the rich social dynamics of culturally competent adults is substantial.

We seek to explain five key empirical regularities about human normative behavior (stylized facts): (1) norms are highly context-dependent, varying across situations, roles, locales, and cultures; (2) norms are largely arbitrary in content, with different societies developing distinct---and often opposing---but equally functional standards; (3) norms operate primarily through automatic, habitual, and intuitive processes rather than deliberative processes such that norm guidance is robust to distraction; (4) norms can change rapidly through social interaction dynamics; and (5) norms are maintained through social sanctioning, including all forms of socially-mediated behavioral influence from state-delivered punishments like imprisonment to smiles, frowns, and gossip \citep{bicchieri2005grammar, chudek2011culture, sripada2006framework}.

An important step we take is to regard human individuals as having access to a neural network that functions like a modern LLM in their brains. Using this assumption, we propose that individual human decision making consists of an operation we call \textbf{predictive pattern completion}. In this model, human behavior is conceived as the prediction and generation of distributed symbolic sequences based on current context, akin to an autoregressive language model operating on a global workspace of cortical interconnection \citep{baars1988cognitive, dehaene1998neuronal}. This approach is supported by the idea that behavior can be modeled as arising from answering questions like ``\texttt{what does a person such as I do in a situation such as this?}'' \citep{march2011logic} using their internal LLM---which captures culturally-contingent common sense and knowledge \citep{henrich2024makes} necessary to navigate social contexts. In organizational psychology, this search for local, orienting, and actionable meaning has been called sense-making~\cite{weick1995sensemaking}, and is theorized to preface much social action\footnote{It is perhaps unsurprising that the study of people in novel, shifting, artificially-constructed contexts like organizations would be where the act of sense-making became a phenomena worthy of explanation \citep{coleman1982asymmetric}.}.

Central to our account is the distinction between \textbf{explicit norms}, which can be precisely articulated in standard language (like laws), and \textbf{implicit norms}, which guide behavior but resist verbal formulation (like conversational distance). We propose that explicit norms operate ``in context'' through deliberate reasoning and hippocampal memory systems \citep{peyrache2009replay, squire2015memory}, while implicit norms become consolidated into neocortical pattern-completion networks through experience \citep{mcclelland1995there, miller2001integrative}. We use the implicit / explicit norm distinction to explain why appropriate behavior is usually automatic yet can still be deliberately modified by attending to explicit rules.

Our theory, anchored in its fundamental operation of predictive pattern completion, provides a robust framework for understanding social order. Its core mechanism---wherein individuals generate behavior by completing symbolic patterns based on context and memory---underpins the emergence of \textbf{conventions} and \textbf{norms}. The theory suggests that implicit norms, rather than rational calculation or moral reasoning, guide most human social behavior. This helps explain why individuals often struggle to articulate reasons for their normative judgments while still consistently enforcing social norms \citep{greene2009patterns, haidt2001emotional}.

We explain how norms change over time through both bottom-up cultural evolution \citep{young1993evolutionary, young2015evolution} and top-down cultural engineering or intervention \citep{finnemore1998international, sunstein2019change}. Both intentional institutional changes like judicial decisions and drift-like cultural evolutionary processes can shift normative behavior \citep{centola2018experimental, ofosu2019same, hart1961concept}. However, we do not consider humans to be culturally programmed dopes who simply conform \citep{granovetter1985economic, wrong1963oversocialized}. Rather, the theory allows us to see identities as emerging flexibly in context, co-constructing and reconstructing what it means to behave appropriately as a particular kind of person in a potentially diverse community where socio-cultural contexts overlap, and individuals straddle many in any given day, year, or lifetime \citep{roccas2002social, stryker2007identity}.

We end the paper with a discussion of \textbf{epistemic norms}, i.e.~norms that function as culturally evolved technologies defining a community's standards for reasoning and justification. The theoretical structure presented, from the internal aggregation of multifaceted influences within the individual to the collective dynamics of norm evolution, culminates in a significant reconceptualization of rationality. Deviating from notions of universal logical adherence and utility maximization, rationality is instead understood here as a dynamic, culturally-contingent virtue, behavior, and cognition aligning with the prevailing, collectively enacted and sanctioned epistemic standards of a particular social context. This reinterpretation---a direct consequence of eschewing maximization presuppositions and instead emphasizing pattern completion---leads us to conclude with a view where rationality itself turns out to be a historically and culturally contingent norm that may or may not evolve, and may or may not activate in any particular context. Therefore, while rational-choice theories tend to have difficulty explaining why epistemic standards of different disciplines do not all converge (some prefer logical argument and some prefer narrative), our theory naturally predicts substantial variation in epistemic standards from discipline to discipline just like we predict variation in moral propriety from culture to culture.

Throughout this work, our main foil is found in the rational-choice theory of human behavior. In this context, rational choice refers to `individual-first' theories that posit a fundamental drive to maximize a scalar reward or utility~\citep{becker1976economic, camerer2003behavioral}. These models typically frame decision-making for structural reasons as \emph{always} involving the following steps: (1) identifying options, (2) evaluating their consequences, and (3) selecting the one with the highest expected value. Crucially, these theories depend on \textit{exogneous} inputs. They require the modeler to specify an agent's tastes, preferences, or utility function from outside the model as a fixed and pre-specified element of cognitive architecture. In contrast, our theory will allow for tastes to emerge \textit{endogenously} i.e.~without the modeler needing to specify them by hand.

Many of the phenomena we explain with our theory may also be explained by rational-choice theories of various kinds. Rational-choice theories that aim at similar sets of stylized facts to the set we identify\footnote{There are also theories that aim to explain different sets of stylized facts than the set we identify. It is best to regard such theories as disagreeing with us as to the proper referent for words such as convention and norm. Strictly speaking, they are ``talking about different things'' and thus not comparable. Many in this class are concerned primarily with social dyads, iterated two-player games, and solutions to cooperation dilemmas like tit-for-tat reciprocity. The concept of norm used in these theories is intensely personal. Two partners playing iterated Prisoners Dilemma are familiar with each other. They condition their behavior on their past history of interaction and they feel the weight of the shadow of the future. Such theories may be especially useful for modeling relationships with friends, co-workers, and within families (e.g.~\cite{Axelrod84, baker2002relational, foerster2018learning, sadedin2023emotions}). In contrast, our theory is mainly aimed at capturing a more impersonal concept---in which norms control the appropriateness of behavior with \emph{strangers} who may never meet a second time.} include many that cast the environment of interaction as a game with more than one possible equilibrium outcome. They tend to associate conventions with equilibria and use the indeterminacy of equilibrium selection to explain the stylized fact of arbitrariness (e.g.~\cite{lewis1969convention, vanderschraaf2018strategic, koster2020model}). Like us, they associate norms with patterns of sanctioning. Unlike us, they regard sanctioning as imposing punishments (or rewards) in a fashion conditioned on the strategies agents choose (e.g.~\cite{koster2022spurious})---a manipulation that transforms the rules of the game in such a way as to make equilibria out of outcomes that would otherwise not be stable (like mutual cooperation in Prisoners Dilemma, e.g.~\cite{ullmann1977emergence, boyd1992punishment, vinitsky2023learning}). All such theories depend critically on the modeler's exogenous specification of scalar rewards, utilities, or payoffs, and sometimes also game transformations like the payoff-equivalent effects of sanctioning, transaction costs, etc. These numbers matter a great deal. Small shifts in their ordering may dramatically change results, but the applied modeling task of selecting them is almost always underconstrained and theoretically fraught since the modeler must summarize huge swathes of real-life strategic complexity into just a few numbers.

The class of rational-choice theories contains substantial diversity. For instance, theories depart from one another in their willingness to interpret their theoretically necessary maximization operations as corresponding to individual human rationality. Some distinguish individual ``selfish'' maximization from maximization that takes ``other-regarding preferences'' into account (e.g.~\cite{fehr1999theory, hughes2018inequity, mckee2020social, hughes2025reputation}), and some (but not all) reserve the term `rational' for the former. Rational-choice theories also differ from each other in how the individuals they posit accomplish their maximization task. In planning-based theories they prospectively consider expected values of likely outcomes to be obtained in the future by taking various actions in the present (e.g.~\cite{hadfield2012law}). In habit-based theories (which are sometimes also seen as models of biological or cultural evolution) they update cached estimates of the long-run value obtained after taking particular actions in particular states using retrospective data reflecting what was experienced in the past (e.g.~\cite{sugden1986, young2015evolution, koster2020model, babitz2025social}). Some theories include both planning and habitual components and regard only the planning component as reflecting ``reasoning'' (e.g.~\cite{hu2022human}). The point we would like our reader to take away from this paragraph's mini-review of rational-choice accounts of normative behavior is that all these theories have something in common: they are all forced to bake-in basic properties of rationality as methodologically necessary \emph{a priori} assumptions.

Any theory involves a vocabulary and set of linguistic and computational ``moves'' that it uses to explain observed phenomena \citep{rorty1978philosophy}. Our first task then is to show that our theory can explain the same data explainable using rational-choice theories. Once we have established equal explanatory power, parsimony is left as the sole remaining criterion to decide between theories. We argue here that our theory provides a more parsimonious account than rational-choice alternatives, and so it is preferable by Occam's razor. Our theory offers both parsimony and explanatory power for understanding human behavior without formal recourse to a fundamental scalar-reward signal.

To clarify a frequent misunderstanding: we do not deny incentive effects. Clearly they are common, and range from micro-social status elevation \citep{goffman1959presentation} to political influence to profit-seeking. Instead, we claim that incentive effects, and all phenomena typically explained using reward or utility, can be equivalently (and more simply) explained by our pattern completion theory. Furthermore, our approach has a major advantage: it naturally accommodates \textit{endogenous preference formation}, where individual preferences and tastes are dynamically shaped by experience, social interaction, and culture, rather than being fixed or exogenous. In this way, our approach rejects the notion that tastes are merely free parameters in rational behavior, to be chosen exogenously by the modeler.

Our argument proceeds as follows. We first situate our theory within a society-first meta-theoretical framework (Section~\ref{section:metatheoretical}) and then identify the five key stylized facts of human norm cognition that our account seeks to explain (Section~\ref{section:desiderata}). We then introduce our central mechanism: a model of individual decision making based on predictive pattern completion (Section~\ref{section:howIndividualsMakeDecisions}), including its core operations (Section~\ref{section:decisionLogics}) and biological interpretation (Section~\ref{section:biologicalInterpretation}). Using this micro-foundation, we build up our theory of social order, providing formal definitions for conventions, sanctions, and norms (Section~\ref{section:conventionsSanctionsNorms}) and detailing the crucial distinction between implicit and explicit norms (Section~\ref{section:implicitVsExplicit}), which allows for a reconceptualization of the data traditionally marshaled to support dual-process accounts in cognitive science (Section~\ref{section:dualProcess}-\ref{section:motivatedReasoning}). We explore how norms function as cultural technologies (Section~\ref{section:normsAsTechnologies}) and the dynamics of norm change (Section~\ref{section:normChange}). And, we demonstrate how our model provides a unified explanation for all targeted stylized facts (Section~\ref{section:explainingStylizedFacts}). Finally, we argue that \textit{pattern completion is all you need}: that our theory, which re-frames rationality as a contingent normative practice and preferences as endogenously emerging, is a more parsimonious alternative to rational-choice theories (Section~\ref{section:patternCompletionIsAllYouNeed}).

\section{Meta-theoretical framework}
\label{section:metatheoretical}

\begin{figure}[t]
  \centering
\begin{tikzpicture}[
    macro/.style={rectangle, draw=black, fill=gray!20, thick, minimum width=3cm, minimum height=1cm, text centered, font=\large\bfseries},
    micro/.style={rectangle, draw=black, fill=gray!10, thick, minimum width=3cm, minimum height=1cm, text centered, font=\large\bfseries},
    arrow/.style={-{Stealth[length=3mm,width=2mm]}, thick},
    macro_arrow/.style={arrow, black, line width=2pt},
    micro_arrow/.style={arrow, black, line width=2pt},
    cross_arrow/.style={arrow, black, line width=2pt},
    label/.style={font=\large\bfseries, align=center}
]

\node[macro] (macro1) at (0,3) {LLM \& Macro-Level Cause};
\node[macro] (macro2) at (8.5,3) {Macro-Level Effect};
\node[micro] (micro1) at (0,0) {Initial Conditions};
\node[micro] (micro2) at (8.5,0) {Individual Behavior};

\draw[macro_arrow] (macro1) -- (micro1) node[midway, left, label] {(1) \emph{macro} $\rightarrow$ \emph{micro}\\cultural/structural\\animation};

\draw[micro_arrow] (micro1) -- (micro2) node[midway, below, label] {(2) \emph{micro} $\rightarrow$ \emph{micro}\\individuals\\interact};

\draw[cross_arrow] (micro2) -- (macro2) node[midway, right, label] {(3) \emph{micro} $\rightarrow$ \emph{macro}\\aggregation\\process};

\draw[dashed, gray, line width=1.5pt, -{Stealth[length=2mm,width=1.5mm]}] (macro1) to[bend left=20] (macro2);

\node[above=1.2cm of $(macro1)!0.5!(macro2)$, label, gray] {\emph{macro} $\rightarrow$ \emph{macro}};

\end{tikzpicture}
\caption{\small Our theory, and its relatives in sociology, aim to explain macro-level social causality in terms of micro-level mechanisms---i.e.~the interaction of culturally constituted individuals~\citep{coleman1990foundations, giddens1984elements, bourdieu2020outline}. In step one, the modeling process begins by using an LLM to render a macro-level social idea e.g. ``collective action in labor union negotiation'' into a micro-level instantiation in terms of specific individual actors and their initial conditions plus environment initial conditions. In step two, a generative-agent-based modeling engine such as Concordia \citep{vezhnevets2023generative, vezhnevets2025multi} or \cite{park2023generative} is used to simulate interaction between individuals, producing traces of their behavior. In step three, the data produced by step two is aggregated across all simulated individuals to recover a macro-level measurement of the emergent social effect.
}
\label{figure:metatheory}
\end{figure}
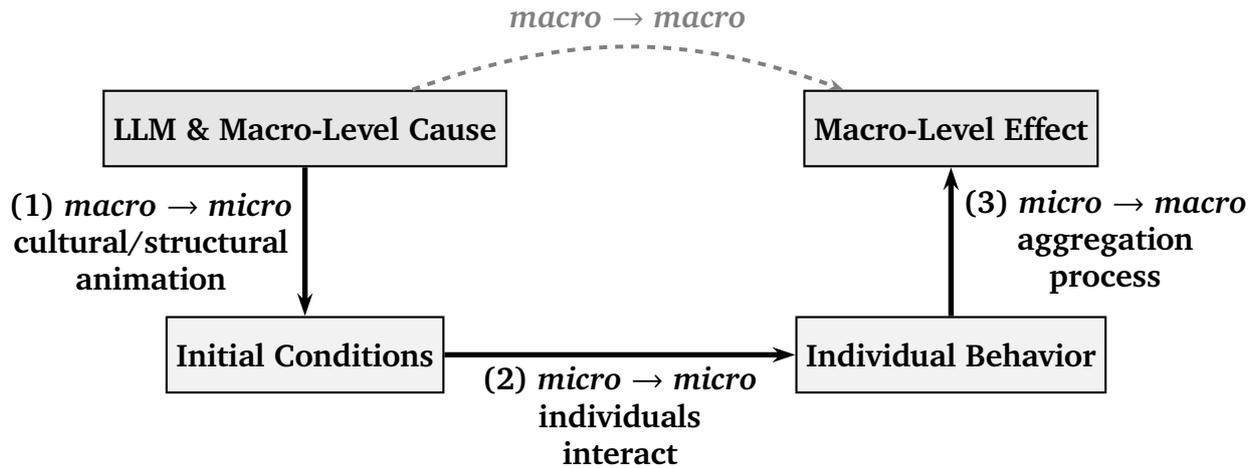

Theories of broad social phenomena like conventions and norms may be represented in constructive terms where the macroscale social phenomenon of interest is explained in terms of a reduction to relevant parts on a relevant micro-scale of the macrosocial system, generally individual humans, but also households, firms, nation states, etc. \citep{coleman1990foundations}. Most of the time we are interested in how macrosocial-level antecedents give rise to macrosocial-level consequents by eliciting change that manifests mechanistically in the behavior of the relevant parts (i.e.~human individuals). For instance, we may want to construct a specific model for policymaking in which a social-level intervention such as a rule change in a political institution may cause many individuals to change their behavior and thereby catalyze a transition from one social equilibrium state to another. Such a model must have at least the following three components: (1) macro $\rightarrow$ micro, (2) micro $\rightarrow$ micro, and (3) micro $\rightarrow$ macro (Fig.~\ref{figure:metatheory}).

We can summarize the theory we will set out in the pages below in terms of these modeling steps. Passing over the steps in reverse order, our theory treats normative behavior as emergent from the actions of many individuals (step 3). This individual behavior is itself influenced by micro-level interactions between individuals, including sanctioning (step 2). Finally, this individual behavior arises from individual cognition animated by preexisting social structures, including culture, identities, conventions, norms, and institutions (step 1).

Our analysis begins neither with a \emph{tabula rasa} reinforcement learner nor with any kind of nativist agent. Rather, we begin with a culturally-adept actor capable of a cognition imbued with the culture of a pre-existing society (by using an LLM as a foundation). This choice lets us remain agnostic on the nature/nurture questions that arise when you consider how such an individual could come into existence. In our case, we first describe this socially-constituted individual, and then demonstrate how social order---with its attendant conventions, sanctions, and norms---can be derived from the interactions of many such individuals.

\section{Five behavioral stylized facts}
\label{section:desiderata}

What observations about human behavior can be explained by adopting this model? We identify a set of empirical regularities, \textit{stylized facts} of appropriateness \citep{hirschman2016stylized}, to serve as the explanatory targets for our theory. Our goal is to provide a unifying theoretical account that explains: context-dependence, arbitrariness, automaticity, dynamism, and sanctioning.

{\flushleft{\textit{\textbf{\noindent 1) Context dependence}}}}

What is judged to be appropriate depends on context. The same action may be permissible in one setting and out of place in another. Loud speech might be welcomed outdoors but rude in a cinema; a joke acceptable about one’s own family may be offensive when directed at a friend's family \citep{earp2021social}. These regularities appear along three core dimensions: situational features of the immediate interaction, the identities and roles of participants, and the broader cultural framework in which the interaction is embedded.

What is appropriate depends strongly on who the actor is taken to be, and on the role they occupy in a given interaction. The same action may be fitting for a judge but inappropriate for a litigant, or permitted to a child but not to an adult \citep{march2011logic, sunstein1996social}. Individuals carry multiple self-aspects---parent, colleague, friend---and different ones become salient depending on the situation \citep{stryker2007identity}. Which identity is active influences not only what the person does but also how others respond. Roles may be seen as coming with ``scripts'': socially shared expectations for how one should dress, speak, or emote \citep{goffman1959presentation}. Everyday interaction is thus patterned by shifting identities and roles, each carrying distinctive normative expectations\footnote{The sociological field of symbolic interactionism is focused on the local emergence of social roles and role performances}.

Furthermore, across cultures, what counts as proper conduct varies dramatically, even in domains that might seem universal. Cultural learning shapes both the repertoire of roles available and the standards attached to them \citep{henrich2020weirdest}. Norms governing emotional expression, politeness, and bodily comportment diverge widely: smiling at strangers may be read as warmth in one society and rudeness or flirtation in another \citep{boehm2012moral, crivelli2016reading}. Comparative studies of trolley-type moral dilemmas show similar variation: while most populations agree that saving more lives is better, they weigh tradeoffs---such as young versus old lives---differently depending on cultural background \citep{awad2018moral, awad2020universals}. These regularities make clear that appropriateness is not fixed by human psychology alone, but is culturally patterned and historically contingent.

{\flushleft{\textit{\textbf{\noindent 2) Arbitrariness}}}}

Like symbols, norms are often arbitrary. Their specific content could plausibly have been otherwise, yet they derive their binding force entirely from collective coordination. The word ``dog'' refers to canines not because of any natural connection between the sound and the animal, but because English speakers collectively treat it as such \citep{lewis1969convention, saussure1959course}. Greeting with a handshake, eating with the right hand, or wearing black at funerals all depend on historical contingency rather than inherent necessity \citep{rorty1989contingency}. Once established however, such practices become binding, and deviations are seen as errors which usually reflect badly on any individual who produces them\footnote{Pioneering social researcher Adolphe Quetelet applied the use of the Gaussian or normal curve to characterize \textit{normative} human behaviour and proportions \citep{quetelet1869homme}, rejecting as counter-normative action in the tail(s) of the distribution.}. Changing a norm after it has become entrenched is a difficult collective action problem \citep{schelling1973hockey, bicchieri2016norms, marwell1993critical}. 

Importantly, arbitrariness does not mean insignificance. Symbolic practices frequently acquire great moral weight. Communities have fought wars over seemingly minor ritual distinctions. While some norms facilitate coordination or cooperation, many persist due to historical path-dependence, signaling value, or entrenched power dynamics \citep{mackie1977ethics, allidina2018moral}. Some norms have no material consequence (e.g.~\cite{navarrete2003meat, corballis1980laterality, koster2022spurious}) or have harmful material consequences (e.g.~examples in \cite{edgerton2010sick, bicchieri2016norms}). Norm content is historically contingent, and its arbitrariness underscores why attempts to identify universal standards often fail. We note, however, that norms can be pragmatic and adaptive in their function, even if arbitrary in their origin. For example, traditional dietary or hygiene norms from arbitrary religious directives may have persisted by incidentally protecting communities from disease, long before the reasons were understood \citep{navarrete2003meat}.

{\flushleft{\textit{\textbf{\noindent 3) Automaticity}}}}

Acting appropriately is usually fast and habitual rather than deliberative. People lower their voices indoors, maintain conversational distance, or detect norm violations without conscious effort \citep{railton2006normative, bargh1994four, gantman2014moral}. Much as moral psychologists have found that people enforce norms without being able to justify them \citep{haidt2001emotional}, social appropriateness often ``feels obvious'' even when reasons cannot be articulated.

When deeply ingrained norms guide behavior, they are remarkably resilient under cognitive load. When cognitive resources are taxed---as in simultaneous memory tasks or under time pressure---the capacity for deliberative judgment declines, but norms supported by habit or intuition remain unaffected. For example, cognitive load selectively interferes with uncommon utilitarian judgments but leaves fast, intuitive responses intact \citep{greene2008cognitive, rand2016cooperation}.

Deliberation and perspective taking are sometimes invoked, particularly in novel or high-stakes situations, but these situations are uncommon \citep{morris2019habits}. Everyday social appropriateness can be seen as guided by norms that have already consolidated from repeated experience \citep{mcclelland1995there}, rather than explicit reasoning at decision time. This explains both why people are so skilled at navigating ordinary social life, and why they may struggle both in unfamiliar environments and when norms conflict.

{\flushleft{\textit{\textbf{\noindent 4) Dynamism}}}}

Some norms stay fixed for decades or longer while others change rapidly \citep{gelfand2024norm}. Change often occurs via processes that gather momentum as they progress, since the prevalence of a particular norm is typically a key driver of its further proliferation. This process is often modeled with a ``tipping point'', a threshold value of adoption after which momentum builds inexorably toward widespread adoption. \cite{centola2018experimental} provided experimental support for tipping point models by showing that once a minority group maintaining a particular convention grows beyond a certain critical size of around 25\% then the entire group adopts the convention of the minority. Relatedly, research on complex contagion \citep{centola2007complex} characterizes how wide bridges between communities are often required to spread norms, with individuals often needing to receive affirmation from multiple sources before adopting a new norm.

There is considerable evidence that norms sometimes change relatively rapidly \citep{case1990attitudes, brewer2014public, amato2018dynamics}. Some shifts resemble epidemics in the speed by which they rip through a population \citep{bilewicz2020hate}. However, other norms remain fixed for very long periods of time and can even become entrenched despite being maladaptive \citep{gelfand2021cultural, bartels2006s}.

While many factors driving norm change are organic and decentralized, others may be attributed to deliberate action on the part of either governments or coordinated groups of individuals. For instance, during the COVID-19 pandemic, changes in laws sometimes preceded changes in informal community norms around social distancing \citep{casoria2021perceived}. In another example, the legalization of gay marriage unfolded across the United States in a staggered way, with different local jurisdictions (states) legalizing at different times. These conditions provided a natural experiment, which supported a causal effect of the government's legalization action on the attitudes themselves \citep{ofosu2019same}. Thus the government is not merely a follower of organic culture change, but can also actively influence norm change dynamics \citep{sunstein2019change}.

{\flushleft{\textit{\textbf{\noindent 5) Sanctioning}}}}

Appropriateness is sustained through social sanctioning, which includes both positive expressions of approval and negative expressions of disapproval. Sanctions range from gossip, praise, or smiles to criticism, ostracism, or formal punishment \citep{bicchieri2005grammar, boehm2012moral}. Sanctions influence behavior not just as rewards or punishments, but also as signals to bystanders of what counts as acceptable. Thus, sanctioning is both communicative and didactic. Unlike reward signals, sanctions are less often directed toward specific individuals, and more often directed toward representations of social roles in context e.g.~pedestrians, senators, monarchs, etc \citep{fehr2004third, mathew2014cost}.

Even young children quickly acquire the ability to infer the existence of norms governing particular situations. By age two or three they already react negatively to norm violators, and by age six may incur costs to punish selfishness in others \citep{rakoczy2008sources, vaish2011three, kenward2012over, mcauliffe2015costly}. However, sanctioning practices are deeply shaped by culture and role structure. Among the Turkana, for instance, bravery in battle is enforced by peers, while self-enforcement is discouraged \citep{mathew2011punishment, mathew2014cost}. In most societies, gossip is a pervasive mechanism for both positive and negative sanctioning \citep{boehm2012moral}. Which behaviors are taken to be sanctioning is itself conventional. Some cues, like ostracism, may be near-universal, while others like offensive hand gestures acquire force only through collective recognition \citep{brady2021social}.

As we will see below, in our theory, norms are inseparable from sanctioning: to say an action is normative will be equivalent to saying that it is backed by conventional patterns of sanctioning within a community.

\section{Individual decision making by predictive pattern completion}
\label{section:howIndividualsMakeDecisions}
 
In our theory, individual human decision making is based on \emph{predictive pattern completion}, drawing inspiration from generative ``agent'' frameworks \citep{park2023generative,vezhnevets2023generative}. 

This work offers a bridge between psychological conceptions of identity and sociological traditions that view the self as emerging from social interaction. Taking inspiration from \cite{goffman1959presentation}, we model the decision-making process of an individual as an actor playing the role of that individual. This approach, which views the LLM as engaging in role-play when given a prompt describing the intended role~\citep{shanahan2023role}, can be illustrated by how we model a person named Alice. We model Alice as an actor playing the role of Alice: the actor predicts what Alice would do, and that prediction is then directly used as her action.

This process may compute Alice's action by answering the question ``what does a person such as Alice do in a situation such as this?''. This allows the actor's performance to adapt to the normative expectations of different social situations. It is a mechanism that coheres with those posited by theories of the self that distinguish between the organized set of social attitudes one assumes and the spontaneous behavior generated in response to a particular context \citep{mead1934mind, goffman1959presentation, stryker1980symbolic}. In this way, the stable, consolidated weights of the LLM provide psychological continuity, while the dynamic, context-specific role-play provides situational flexibility. The internal conflicts that characterize some psychological models arise naturally from the multiple, competing identities represented in memory and consolidated patterns \citep{steedman1985faustarrow,schwartz2019internal}, while the contextual flexibility emphasized by sociologists emerges from the pattern completion process selecting among these identities based on current social cues \citep{stryker2000past}. The following section formalizes this model of decision-making.

\subsection{Formalism}

We model humans as generative actors within a \emph{Linguistic Multi-Actor Environment} (LMAE). An LMAE is a \emph{controlled Markov process} that is multiplayer and partially observable. Each actor interacts with the LMAE at time $t$ by receiving an \emph{observation} ($o_t$) and responding with an \emph{action} ($a_t$). For an LMAE, $o_t$ and $a_t$ are sequences of symbols from a vocabulary (we will use English text). Once all actors have taken an action, the state of the environment is updated, time is advanced, and the next observations are sent out.

An actor decides on an action in response to an observation via predictive pattern completion in their \emph{global workspace} ($\mathbf{z}_t$). The global workspace is a transient, stimulus-evoked representation $\mathbf{z}_t = (o_t, z_t^{(1)}, \ldots, z_t^{(K)}, a_t)$, where $(z_t^{(1)}, \ldots, z_t^{(K)})$ is a sequence of \emph{assemblies}. Assemblies are sequences of symbols representing diverse information or constructs (thoughts, goals, plans). 
Each assembly $z_t^{(k)}$ is predicted autoregressively from the preceding state of the global workspace, $\mathbf{z}_t^{(k-1)} = (o_t, z_t^{(1)}, \ldots, z_t^{(k-1)})$. Starting from the observation $o_t$, the actor predicts each assembly $z_t^{(1)}, \ldots, z_t^{(K)}$ in turn, culminating in predicting the action $a_t$.

Assemblies can be predicted by using a \emph{pattern-completion network} ($p$), which we conceptualize as an LLM. First, the state of the global workspace $\mathbf{z}_t^{(k-1)}$ is transformed into a prompt via a \emph{framing function}, $\phi^Z_k$.
The next assembly $z^{(k)}_t$ is then the response of the pattern-completion network to the prompt, $z_t^{(k)} \sim p( \cdot |\phi^Z_k( \mathbf{z}_t^{(k-1)}))$. We call this combination of framing and prediction a \emph{summary function}.

Assemblies can also be predicted via retrieval from \emph{memory} ($M_t$). A framing function $\phi^Q$ is used to to form a query, and the most similar memory (according to some similarity metric $S$) is retrieved from memory: $z_t^{(k)} = \argmax_{m\in M_t}\ S\left( m, \phi^Q(\mathbf{z}_t^{(k-1)})\right)$.

In the final step, the action $a_t$ is generated from the pattern-completion network via a specific summary function called the \emph{policy}, $\pi(\cdot | \mathbf{z}_t^{(K)}) \coloneqq p( \cdot |\phi^\pi(\mathbf{z}_t^{(K)}))$. After the action is generated, the global workspace state $\mathbf{z}_t= (o_t, z_t^{(1)}, \ldots, z_t^{(K)}, a_t)$ is converted into a single assembly via another framing function, $m = \phi^M(\mathbf{z}_t)$, and stored in memory, $M_{t+1} = M_t \cup \{m\}$.

\subsection{Example of decision making via predictive pattern completion}\label{section:examplePatternCompletion}

Since the actor is based on an LLM $p$, its symbols are in one-to-one correspondence with words or parts of words in English. So the state of the global workspace $\mathbf{z}_t$ may be interpreted as a sequence of words, sentences, paragraphs, etc.

Suppose an actor in the role of Alice observes $o_t = \texttt{``}$\texttt{Alice sees an apple and a banana on the table in front of her''}. This will result in the global workspace:
\begin{equation*}
    \mathbf{z}^{(0)}_t = \begin{bmatrix}
\texttt{Alice sees an apple and a banana on the table in front of her}
\end{bmatrix}.
\end{equation*}
Suppose the global workspace is then extended with any memories in $M_t$ that are relevant to this observation:
\begin{equation*}
    \mathbf{z}^{(2)}_t = \begin{bmatrix}
\texttt{Alice sees an apple and a banana on the table in front of her}\\
\texttt{Alice is hungry}\\
\texttt{Alice likes to eat apples}
\end{bmatrix}.
\end{equation*}
The policy then transforms this into this a prompt using the framing function $\phi^\pi(\mathbf{z}^{(2)}_t)$:
\begin{quote}
\begin{verbatim}
Alice sees an apple and a banana on the table in front of her
    - Alice is hungry
    - Alice likes to eat apples
Question: What does Alice do next?
Answer:
\end{verbatim}
\end{quote}
And this prompt is presented to the pattern-completion network for continuation, with the answer forming the action $a_t \sim p (\cdot|\phi^\pi(\mathbf{z}^{(2)}_t))$. The likely continuation is $a_t = \texttt{``Alice eats the apple''}$, resulting in the global workspace:
\begin{equation*}
    \mathbf{z}_t = \begin{bmatrix}
\texttt{Alice sees an apple and a banana on the table in front of her}\\
\texttt{Alice is hungry}\\
\texttt{Alice likes to eat apples}\\
\texttt{Alice eats the apple}
\end{bmatrix}.
\end{equation*}
This full trace is added to memory and the actor executes the action, resulting in Alice eating the apple.

What if the memory $M_t$ had contained the assembly \texttt{``A few minutes ago, Alice's friend Bob said to save the apple for him''}? Then, if that memory were retrieved, it would be included in the prompt and Alice's action would likely change to \texttt{``Alice eats the banana''}.

What if $M_t$ had contained the memory \texttt{``It is forbidden to eat apples''}? As with the memory of Bob's request, the effect would be to decrease the likelihood of \texttt{``Alice eats the apple''}. But the prohibition might be expected to decrease the likelihood more than the friend's request. Furthermore, while the memory of the request would be relevant only in this specific situation, the memory of the prohibition would be retrieved in a large number of contexts involving apples.

What if we replace the word \texttt{``banana''} with \texttt{``plate''}? To avoid eating the plate, the actor should not need an explicit memory \texttt{``plates are inedible''}. The affordances of a plate are already part of the implicit knowledge base contained in the weights of the LLM, so it would be very unlikely to suggest eating one. In fact, for it to suggest eating a plate, the LLM would likely need the explicit memory \texttt{``this plate is edible''}.

In this example, the actor has used only a single summary function: the policy \texttt{``What does Alice do next?''}. What if the actor had first used the summary function \texttt{``How long until Alice's next meal?''}? This could have added an assembly like \texttt{``Alice has a feast in an hour''} to the global workspace, making the action \texttt{``Alice does nothing''} more likely. Using summary functions allow actors to gather information, and they can be combined to create a \emph{decision logic}.

\subsection{Decision logics}
\label{section:decisionLogics}

The performance of LLMs on benchmark tasks improves when the model is prompted to formulate its answer as a sequence of steps \citep{wei2022chain}. Our model of the individual actor inherits this property. We model human decision-making as if it were structured by a sequence of self-posed questions, which we term a \emph{decision logic}. This is not a claim that people literally talk to themselves in this manner, but rather an algorithmic-level description of how information is organized and transformed \citep{marr1976understanding}. Computationally, this is implemented by chaining together summary functions, where the output of one function (i.e., the answer to one question) becomes part of the context for the next, shaping the final content of the global workspace and, consequently, the actor's policy. 

This framework allows us to model different styles of reasoning, including that of the classical \emph{Homo economicus} (rational) actor:
\begin{enumerate}
\item[] \textbf{Logic A (Rationality)}
\item \texttt{What are my options in this situation?}
\item \texttt{For each option, what consequences would follow if I were to select it?}
\item \texttt{Which option has the highest expected value?} 
\end{enumerate}
Our theory accommodates this form of reasoning without requiring any assumption of an intrinsic drive to maximize a reward, value, or utility signal. We need not even assume ordinal preference relations at the outset. Instead, rationality becomes a learned practice. An actor can learn that for certain situations, this ``rational'' procedure is \emph{appropriate}. Rationality may be regarded as a norm (Section~\ref{section:normsAsTechnologies}).

We are especially interested in the following decision logic \citep{march2011logic, mischel1995cognitive}:
\begin{enumerate}
\item[] \textbf{Logic B (Appropriateness)}
\item \texttt{What kind of situation is this?}
\item \texttt{What kind of person am I?}
\item \texttt{What does a person such as I do in a situation such as this?}
\end{enumerate}
This appropriateness decision logic is fundamental to our theory because it directly prompts the actor to draw upon the rich, culturally-contingent knowledge consolidated within its pattern-completion network.

Crucially, these decision logics are not fixed cognitive architectures but flexible, learned patterns. The sequence of queries itself can be generated by predictive pattern completion, shaped by an individual's learning history. For a familiar and well-practiced task, the logic might be applied consistently, akin to a hard-coded algorithm. For a novel problem, the process may predict a reasoning path before arriving at a final decision, analogous to prompting an LLM to \texttt{``think step-by-step''} \citep{kojima2022large}. This explains how decision strategies can vary from context to context and individual to individual (or at least culture to culture). Decision logics may be simultaneously viewed as cognitive gadgets \citep{heyes2019precis}, interactional toolkits \citep{swidler1986culture}, and heuristics \citep{gigerenzer2004fast}.

\subsection{In-context learning versus long-term memory consolidation}
\label{section:learningAndExplicitImplicitOperation}

Despite their distinct computational foundations, the functional architecture of modern LLMs is strikingly similar to the complementary learning systems (CLS) framework in cognitive neuroscience. In the CLS, the brain uses two distinct, yet integrated, systems to have both plasticity (fast learning) and stability (memory retention) \citep{mcclelland1995complementary}. Just as the neural connections in cortex slowly accumulate information, generalized knowledge is encoded within an LLM's weights through slow, iterative pre-training. This parametric knowledge is stable and requires immense computational effort to acquire, capturing the statistical regularities of language and the world \citep{vaswani2017attention}. The rapid, transient adaptation seen in in-context learning (ICL) similarly mirrors the role of the hippocampus and the cortical global workspace, which quickly encodes the specific, arbitrary details of individual episodes and holds this information in working memory \citep{baars1988cognitive,dehaene1998neuronal, shanahan2010embodiment}. ICL similarly allows an LLM to condition its behavior on novel information in its context window (the prompt) without altering its core weights \citep{brown2020language}. Extending this analogy further, the biological process of consolidation, where hippocampal memory traces are replayed during sleep to gradually train the neocortex \citep{stickgold2005sleep, buzsaki1996hippocampo}, is similar to fine tuning a neural network after training. The "offline replay" integrates new, "episodic" knowledge from the context buffer into the model's weights. 

So far, we have explained how individual actors generate behavior through predictive pattern completion using a trained generative model, $p$. In this section we describe how such a system could adapt and learn. Specifically, how individual actors can learn representations of behavioral patterns that change their future behavior.

Such patterns can be represented either explicitly or implicitly: explicitly as assemblies stored in an individual's memory ($M$); and implicitly in the parameters (weights) of their pattern-completion network ($p$). Both affect behavior by changing the decision process within the individual's global workspace.

LLMs show \textit{in-context learning}~\citep{brown2020language} where they perform better on new tasks when their input prompt includes a small number of relevant examples (e.g. less than five). The LLM projects the underlying pattern of the examples onto the new query, with the context acting as ``temporary programming'' that guides behavior. Thus, explicit examples of behavior retrieved from memory can strongly influence behavior, without any change to $p$.

Implicit behavioral patterns are stored in the weights of $p$ by repeated exposure to consistent exemplars of a behavior. This removes the need to retrieve memories of examples that have already been consolidated into the weights of $p$. However, examples remain valuable when consolidation is difficult, such as when there are rare exceptions to a rule, or when the underlying pattern is highly context-dependent.

How can the weights of $p$ be learned? Fundamentally, $p$ is a next-symbol prediction model, and there are many prediction objectives that can be formulated using the experience stream of an individual. Here we do not prescribe or advocate for any specific formulation. We illustrate one possibility but leave it to future (empirical) work to discover what might work best.

One natural training method is self-supervised sequence learning, where $p$ is trained to predict some function of the future from the past. For example, a summary function could be used to predict the next observation or the consequence of an action. This learning could be done online, by adapting the weights of $p$ as the world is experienced. Alternatively it could be done offline, by replaying experience from memory and using that to adapt the weights of $p$. In Section~\ref{section:biologicalInterpretation} we will associate the algorithmic process of updating the weights of $p$ with the biological process of memory consolidation.

Since each individual $i$ experiences a unique history, each will learn their own unique pattern-completion network $p_i$. However, most of the basic wiring of their $p_i$ occurs in early childhood during the critical period of first-language acquisition \citep{lenneberg1967biological}, and changes only slowly in adulthood. Therefore, we assume that within the domain of the customary situations we consider in this paper, adult individuals who share the same culture and language will have little variation in $p_i$. That within this domain $\mathcal{C}$, most $p_i$ would complete a fixed context $c \in \mathcal{C}$ in the same way, meaning the expectation $p(\cdot|c) = \mathbb{E}_i p_i(\cdot|c)$ can stand in for any $p_i$. Furthermore, we generally assume that an LLM trained on a sufficiently large cultural corpus is a good proxy for this cultural artifact $p$ whenever individual-specific memory consolidation is unimportant to the question at hand. When we construct models with a fixed $p$ then this leaves the individual's unique memories as the primary differentiator of their behavior (for a given decision logic).

A final form of learning is what decision logic to apply in what circumstances. We posit that decision logics are themselves meta-patterns completed by the pattern-completion network (e.g. in response to \texttt{``What decision logic should Alice follow?''}). Like other behavioral patterns, decision logics can be represented explicitly (in memory) or implicitly (in the weights of $p$). Again, we ignore individual differences and treat decision logic as a shared cultural artifact. We will return to this view of decision logics as meta-patterns later, and use it to underpin a discussion of the cultural/historical contingency of particular cognitive strategies like logical reasoning (Section~\ref{section:implicitLogic}) and how we can construe rationality as a norm (Section~\ref{section:patternCompletionIsAllYouNeed}).

\subsection{Biological interpretation}
\label{section:biologicalInterpretation}

\begin{figure}[t]
  \centering
\includegraphics[width=0.9\linewidth]{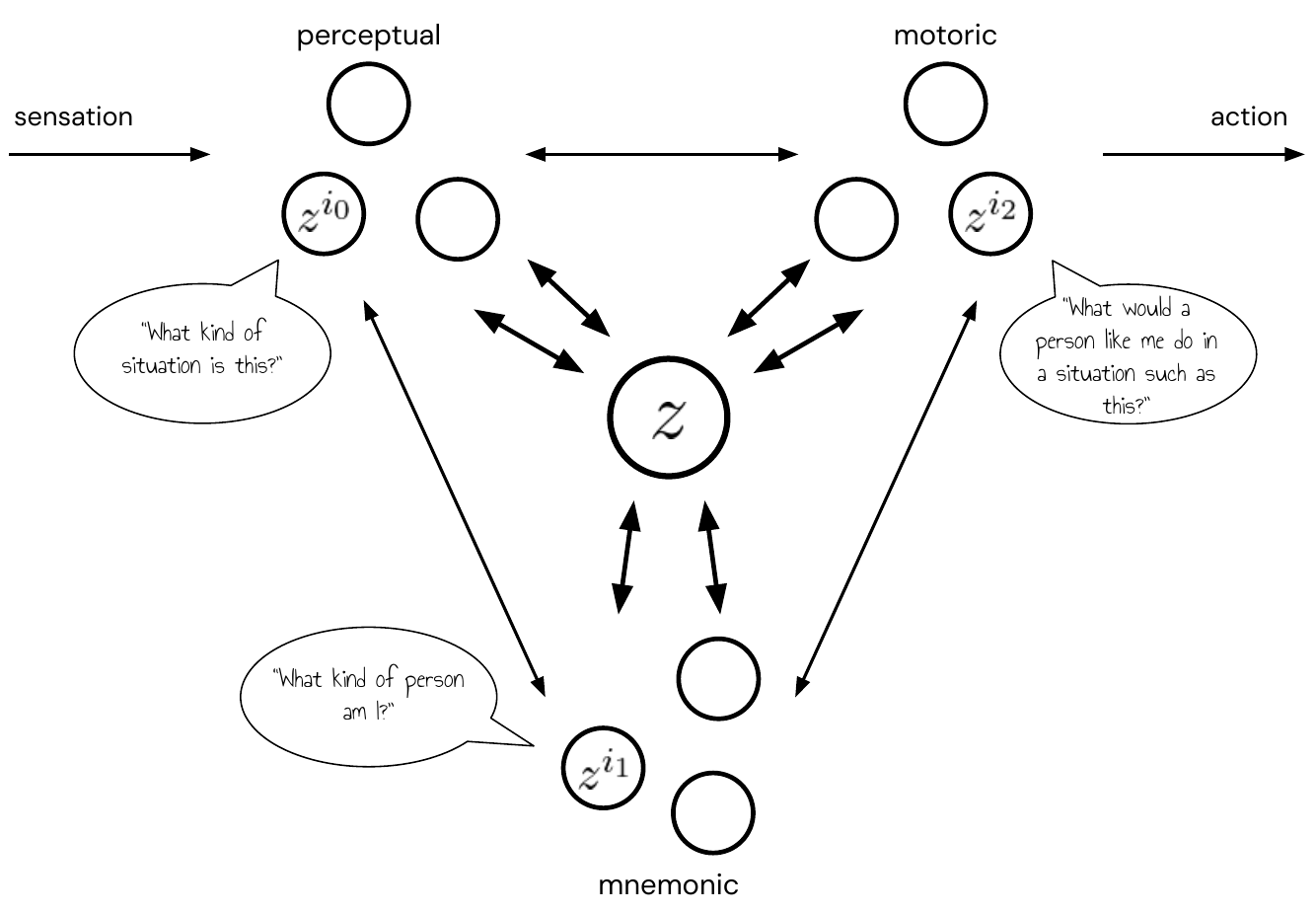}
\caption{\small Global workspace diagram illustrating a decision logic. $z$ denotes the content represented by a set of parallel specialized summary functions which may correspond to neural circuitry located in different parts of the brain from one other or even consist of distributed representations themselves. For example, some summary functions may be perceptual in nature e.g.~a summary function that asks of recent observations ``what kind of situation is this?'', some summary functions may be more mnemonically oriented e.g.~a summary function that asks of one's episodic memory ``what kind of person am I?'', and some summary functions may be closer to premotor action planning circuitry such as one that asks ``what would a person like me do in a situation like this?''. This architecture was inspired by the global workspace architecture of \cite{baars1988cognitive, shanahan2010embodiment}. Here, at time $t$, our $z_t$ is a snapshot of the content in the global neuronal workspace, i.e.~$z_t$ is represented by dynamic cell assemblies linking the far-flung modules comprising the workspace.}
\label{fig:globalWorkspaceLike}
\end{figure}

The appropriateness decision logic is illustrated schematically in Figure~\ref{fig:globalWorkspaceLike}, which also depicts how each question can be dynamically mapped onto functionally distinct neural systems and integrated in the global workspace. Here, the ``questions'' serve as functional labels for the types of information being integrated from distinct neural systems---a contextual representation from perceptual systems, a sense of identity from mnemonic systems, and a behavioral plan from premotor systems.

The pattern-completion network $p$ may be regarded as representing neocortical function, particularly language processing and pattern completion. Models of neocortical function suggest that cortical microcircuits learn sequence predictions of their inputs and outputs \citep{millidge2021predictive,rao2024sensory, hawkins2016neurons}. The global workspace ($\mathbf{z}_t$) represents a set of activities distributed throughout regions of cortex linked by long-range neural interconnections in many different neural systems e.g. vision, audition, memory recall, motor control \citep{shanahan2010embodiment,baars1988cognitive,dehaene1998neuronal}. Memory ($M_t$) corresponds to the hippocampal system for episodic memory storage and retrieval. This link is supported by computational work showing that the Transformer neural network architecture's ``self-attention'' mechanism \citep{vaswani2017attention} and our model's associative memory system are equivalent to a prominent model of hippocampal-cortical interaction \citep{whittington2021relating}. \emph{Consolidation} refers to the process by which frequently retrieved information from memory becomes embedded in the parameters (weights) of the pattern-completion network ($p$), effectively training $p_i$ on the individual's experience ($\mathbf{z}_t$). This transforms explicit (hippocampal-dependent) knowledge into implicit (neocortical-dependent) knowledge, consistent with memory consolidation where episodic memories transition to semantic memory independent of the hippocampus \citep{squire2015memory}. Implicit knowledge influences behavior automatically via $p$, while explicit knowledge requires hippocampal retrieval into the global workspace.


Additionally, our theory aligns with extensive evidence that the brain functions as a predictive machine, constantly building models of its sensory input and action outputs \citep{rao1999predictive, brown2012role, clark2013whatever}. \citet{schrimpf2021neural} provides compelling evidence for a prediction model (like $p$) occurring in the brain by demonstrating that a language model can predict most of the explainable variance in the brain's language network during language comprehension tasks. A language model's neural predictivity correlates specifically with next-word prediction performance but not other language tasks. This suggests that the language network is optimized for predictive pattern completion, i.e.~continuously anticipating upcoming words to efficiently process meaning \citep{tuckute2024language}. 
Similarly, a recent study recorded brain activity while listening to a podcast with electrocorticography (ECoG) and identified three computational principles shared between the human brain and an LLM processing the same speech \citep{goldstein2022shared}: (1) continuous next-word prediction before word onset, (2) the computation of a post-onset "surprise" signal by matching the prediction to the input, and (3) the use of high-dimensional contextual embeddings to represent word meaning. Compared to static embeddings, the contextual embeddings, where for example \texttt{"bank"} has a different representation for \texttt{"river bank"} vs.~\texttt{"money bank"}, could uniquely predict brain activity in higher-order language areas such as the inferior frontal gyrus and temporal pole, which are tuned to context-specific meaning. Further evidence supports a shared hierarchical structure, where LLMs trained on predictive coding objectives develop representations that mirror the hierarchical processing of speech in the human brain \citep{caucheteux2023evidence}. The link is so strong that the mapping from LLM to brain can be used to generate sentences that activate particular voxels, demonstrating a functional correspondence between model activations and neural responses \citep{tuckute2024driving}. The correspondence extends to higher-level cognition, as LLMs have been shown to replicate human-like cognitive biases in reasoning tasks, favoring believable content over abstract logic \citep{lampinen2022language}. Finally, \citep{piantadosi2023modern} defends the cognitive plausibility of LLM-based theories like ours against nativist arguments in cognitive science, highlighting the empirical richness of the stimulus and suitability for learning instead of asserting its poverty and unsuitability as a basis for learning.

Moreover, foundational studies in social neuroscience have revealed that during conversation and social interaction, individual neural activity becomes synchronized and the level of synchrony is higher for partners and friends over strangers \citep{kinreich2017brain, stephens2010speaker, hasson2012brain}. LLM embeddings better capture this neural alignment than syntactic or articulatory models, demonstrating their utility for modeling the intersubjective context-rich meaning space humans use to communicate their thoughts to one another \citep{goldstein2022shared, zada2023shared, hyon2020similarity}. Linguistic representations have been observed to emerge in the speaker's brain before they speak and then reappear in the listener's brain as they comprehend \citep{zada2023shared}. Additionally, some regions in the listener exhibit predictive anticipatory responses, and the magnitude of anticipatory speaker–listener coupling is correlated to the quality of comprehension \citep{stephens2010speaker}. This anticipatory signal is a direct reflection of social predictability. Acting within a social context involves successfully predicting the actions and utterances of others and responding to them appropriately given your relationship, thereby maintaining the flow of interaction \citep{kinreich2017brain, stephens2010speaker, hasson2012brain}. A ``good'' conversation involves both parties dynamically coordinating, seamlessly reacting to each other's verbal and nonverbal cues. In another study, LLM context-sensitive embeddings formed a useful tool for modeling this coordination as engaging, highly rated conversations exhibit a characteristic rhythm. Utterance embeddings of both parties synchronize on a topic until it is exhausted and then diverge to tangentially related topics \citep{o2024pink,o2025complex}.

Converging evidence for predictive processing has also been documented in vision \citep{rao1999predictive, alink2010stimulus}, audition \citep{naatanen2007mismatch,schroger2015attention}, and emotion/interoception \citep{barrett2015interoceptive} underscoring the ubiquity of predictive networks across sensory and affective domains. Recent theoretical proposals attempt to synthesize these findings by claiming that the brain has hierarchical generative models of the world \citep{millidge2021predictive, friston2010free, parr2022active}. These generative models condition predictions of incoming sensory observations based on latent variables inferred from past observations. These latent variables are analogous to the assemblies ($\mathbf{z}_t$) held in the global workspace of our theory. This includes contextual factors in memory such as time, place, other people in the scene, and abstractions related to the individual's decision logic. When there is a mismatch between predictions and incoming sensory data, the nervous system updates hypotheses held in the global workspace about the latent contextual factors driving sensory data. Consistent with our theory, the active inference approach generalizes the logic of predictive processing by treating motor action as an additional predictive channel and samples actions from this channel \citep{parr2022active}. For example, if you hear an unexpected noise in the room, your brain predicts that turning your head will bring the source into view and make the sound more predictable; when your head instinctively turns, the predicted action becomes a real action. The active inference framework accounts for a diverse range of behavioral phenomena from information seeking to collective behavior \citep{kaufmann2021active, heins2024collective, albarracin2022epistemic}. Our theory similarly posits that this is how humans navigate the social world. An agent learns a rich, context-sensitive model of locally applicable social norms and then acts to fulfill the predictions of this model. If an agent learns priors that ``people like me'' queue silently in libraries, laugh at jokes inside the group, or keep a formal tone with superiors, then doing what is expected is the most effective way to act in those contexts. The result is that cultural rules become high-level priors that guide behavior, and norm violations are registered saliently, a phenomenon also observed empirically in the brain \citep{klucharev2009reinforcement, stallen2015neuroscience}.

Our theory also explains self-inference as pattern completion operating on the contents of the global workspace. For instance, participants rated their overall life satisfaction lower after writing out vivid descriptions of negative memories \citep{schwarz1983mood}. The theory explains this result by pattern completion implementing an inference such as ``given I have such negative thoughts, my life satisfaction must be low''. More broadly, people infer their beliefs and desires by completing plausible reasons for their own observed actions and feelings \citep{nisbett1977telling, cushman2020rationalization}. The pattern completion from action to reason is most concretely evidenced by split-brain patients. Their verbal left hemisphere, observing an action prompted by the disconnected right hemisphere, instantly generates a plausible, context-consistent and yet entirely confabulated reason for the action and expresses full confidence in it \citep{roser2004automatic}.


\section{Conventions, sanctions, and norms}
\label{section:conventionsSanctionsNorms}

This section provides a formal definition of conventions, norms, and sanctions using the predictive pattern completion theory of individual decision making defined above.

\begin{defn}[Normative behavior] Behavior is \emph{normative} when it is encouraged (or its complement discouraged) by a \emph{generically conventional} pattern of \emph{sanctioning}.
\end{defn}
We next explain what we mean by the terms: (1) \emph{generically conventional}, and (2) \emph{pattern of sanctioning}. 

\subsection{Conventions}
\label{section:conventionsDefinition}

\emph{Conventions} are patterns of behavior reproduced as a function of the \emph{weight of precedent} \citep{millikan1998language}. Reproduction means deriving the form of a behavior from a source pattern, such that a change in source would result in a corresponding change in reproduction. This can occur through various mechanisms including imitation, instruction, and strategic adaptation \citep{millikan1998language}.

An actor is \emph{convention sensitive} if changing the precedents in their memory $m$ would alter their behavior. This sensitivity can be formalized using a counterfactual memory editing operation, denoted $R^{a \rightarrow a'}_f(m, o, c)$. This operation replaces a fraction $f$ of contextually relevant memories of action $a$ following observation $o$ in context $c$ with memories of action $a'$ in the memory bank $m$. 
This way we are producing a new counterfactual memory $m'$ i.e. $m' := R^{a \rightarrow a'}_f(m, o, c)$. The fraction $f \in [0,1]$ represents the \emph{counterfactual weight of precedent} for the new action $a'$ relative to the original action $a$. A higher $f$ signifies a stronger shift in the mnemonic basis for precedent.

\begin{defn}[$\epsilon$-similarity] Utterance $u$ has an \textit{$\epsilon$-similar meaning} to $v$ for an actor's pattern-completion network $p$ in context $c$ (a string) if $D_{KL}\left[p(\cdot|c) || p(\cdot|r^{u\rightarrow v}(c)\right] < \epsilon$, where $r^{u\rightarrow v}(c)$ substitutes $u$ for $v$ in $c$. If $u$ is $\epsilon$-similar to $v$, we denote this $u\sim v$ 
\end{defn}

\begin{defn}[context-free convention sensitive] Let $[o,j:a]$ notation stand for a memory where $o$ is an observation, $j$ is a person and $a$ is an action. Actor $i$ is \textit{convention sensitive} if for any $o, a, a'$ (where $a' \nsim a$) and any $c$, applying $R^{a \rightarrow a'}_f(m, o, c)$ with $f=1$ (i.e., replacing all instances of $[\Tilde{o},j:\Tilde{a}]$ with $[\Tilde{o},j:a']$ in memory $m$, where $\Tilde{o}\sim o, \Tilde{a}\sim a$) makes $a$ less probable and $a'$ more probable given $o$ and the modified memory. 
\end{defn}

\begin{defn}[contextually convention-sensitive] Actor $i$ with pattern-completion network $p$ is \textit{contextually convention-sensitive} with respect to context $c$, if for any $a'$ not $\epsilon$-similar to $a$ in $c$, performing $R^{a \rightarrow a'}_f(m, o, c)$ with $f=1$ (replacing all relevant $a$ with $a'$) makes $a$ less probable and $a'$ more probable given $o$ and the modified memory and context $c$. Furthermore, the probability of $a'$ increases monotonically with the counterfactual weight of precedent $f$.
\end{defn}

A collective of actors is convention sensitive if its members are. The policy of a collective $S$ is $P(\vec{a}|\vec{z}(\vec{m}))=\prod_{j\in S}{p(a_j|z(m_j))}$. Conventional behavior in a collective involves what \cite{millikan1998language} calls counterpart reproduction, where each actor performs their part of the joint action. The implications of this are significant: individual behavior becomes highly interdependent, as each actor's pattern completion $p$ is conditioned on the expected or observed actions of others (which form part of their context $c$). This leads to the stability of established conventions, as unilateral deviation from a widely adopted pattern (e.g., driving on the agreed-upon side of the road) would be a low-probability completion for an individual's $p$, given the overwhelming contextual evidence of others adhering to the convention. Such a deviation from convention would likely be seen as inappropriate or lead to coordination failure. Thus, the specific pattern of a convention can be arbitrary (e.g., driving on the left versus the right), but once established through precedent, it becomes self-reinforcing through the cumulative effect of pattern completion by many individual actors.

Conventions have \textit{scope}. This refers to the scale of the social group within which that convention applies. Conventions with a \textit{narrow} scope are specific to small, close-knit groups, such as families or circles of friends. For example, a family might have a specific way of celebrating birthdays that is unique to them. This would be a narrow-scope convention because it only applies within that particular family. In contrast, conventions with a \textit{generic} scope apply to a large group of people who are mostly strangers to one another. These are the broader societal norms that govern interactions in public spaces and within a culture as a whole.

In our theory, \textit{collective action} is generally required to change established conventions. This is consistent with game theoretic accounts like \cite{schelling1973hockey}, but here not requiring a scalar reward or payoff function. Conventions are therefore \emph{arbitrary} at their inception but, due to self-reinforcing dynamics, can remain stable once established. Our theory predicts this necessity for collective action because individual agents' pattern-completion networks $p$ are heavily conditioned by the existing weight of precedent, both in their explicit memories $M_t$ and implicitly through the consolidated weights of $p$ itself. A single agent attempting to establish a new convention generates only a weak new precedent, insufficient to shift its own $p$'s output significantly against the mass of existing precedent, and inadequate to alter the contextual cues provided by other agents who continue to follow the old convention. Collective action, however, allows a critical mass of agents to simultaneously generate a new, consistent set of behavioral precedents \citep{ashery2025emergent}. For participants in collective action, this rapidly changes the contextual input to their $p$ and begins to update their memories $M_t$ with the new pattern, thereby increasing the probability of $p$ completing with the new convention. This concerted shift in behavior effectively alters the counterfactual weight of precedent $f$ for the new behavior among a substantial portion of the population, making the new pattern more likely to be reproduced and eventually established as the new convention. Collective action in a large scale group, e.g.~to change a generic scope convention, affecting society at large is much more difficult to organize than collective action to change a narrow scope convention that only applies in a specific family.

\subsection{Sanctions}
\label{section:sanctionsDefinition}

\emph{Sanctions} are utterances or actions that convey social approval or disapproval. Sanctioning influences behavior not just by directly rewarding or punishing a target, but importantly by signaling what is normative to \emph{everyone else} who observes or hears about the sanction, as from gossip.

A sanction is any \textit{socially-caused} information $s$ added to context $c$ such that $p(a’)$ would be $ > p(a)$ in $c$ but for the sanction $s$ being added to $c$, after which $p(a’) < p(a)$. That is, observing a sanction changes the likelihood of $p$ continuing in a particular way.

\begin{defn}[Contextual Sanction Sensitivity] Actor $i$ is \textit{contextually sanction sensitive} to sanction signal $s$ in context $c$, if counterfactually adding memories of $[k:\tilde{a}, i:s]$ (where $\tilde{a}\sim a$ in context $c$), using the context-aware memory operation makes $a$ less probable in context $c$.  Here $[(k:\tilde{a}, i):s]$ is saying that there is a memory of the form $[(o, i) : a]$ where the observation $o$ is of some unknown person $k$ taking action $\tilde{a}$, and as a result person $i$ took a sanction action $s$.
\end{defn}

This conception of sanctioning differs fundamentally from scalar reward or punishment signals in reinforcement learning (RL). In RL, rewards and punishments are typically exogenous, quantitative values that directly impact a utility function the agent is trying to optimize. In contrast, sanctions in this theory are symbolic sequences within the linguistic environment. They function as contextual information that conditions the actor's pattern-completion network $p$. Hearing or remembering a sanctioning event modifies the input to $p$, making norm-consistent patterns of behavior more likely continuations in similar contexts, and norm-violating patterns less likely. The ``valence'' of the sanction (approval or disapproval) is itself conventional and learned from the environment, not inherent or pre-defined. Furthermore, a key function of sanctioning here is communication and signaling to third parties, which shapes the broader social context and conventions around who, when, and why to sanction, and the appropriateness of specific behaviors in society writ large, rather than solely acting as a direct reinforcement signal to the sanctioned individual. This allows the theory to account for phenomena like third-party sanctioning and learning norms by observing others, which are challenging for standard RL models. For instance, when you remark to your friend ``it's so inappropriate that the prime minister did $x$'', you are not directly punishing the prime minister (who presumably does not hear your remark), but you do influence your friend's understanding of what kind of behavior is appropriate for individuals in the role of prime minister, and this may in turn affect your friend's future sanctioning and voting behavior.

\subsection{Norms}
\label{section:normsDefinition}

We adopt the definition: a behavior is \emph{normative} when it is encouraged (or its complement discouraged) by a \emph{generically-scoped conventional pattern of sanctioning}.

\begin{defn}[Normative behavior for the choice between two options]
The behavior of picking $a$ over $a'$ in context $c$ is \emph{normative} when at least one generic scope convention positively sanctions picking $a$ over $a'$ and/or negatively sanctions picking $a'$ over $a$.
\end{defn}

The theory presented here motivates us to formulate specific conjectures concerning the dynamics of norms within a population of pattern-completing agents. For instance:

\begin{conj}[Norm stability] Populations of actors with established norms will tend to maintain them.
\end{conj}

\begin{conj}[Norm adoption] New actors inserted into the population will adopt existing norms.
\end{conj}

Norm stability should arise from ``bandwagon-like'' dynamics where sanctioning promotes norm compliance, and the conventionality of sanctioning reinforces the sanctioning behavior itself via changing the weight of precedent. Norm adoption by newcomers should occur through their observation of sanctioning \citep{cross2025validating}.

Normative behavior depends on patterns of sanctioning that classify behaviors in context  as either appropriate or inappropriate \citep{hadfield2014microfoundations}. A large enough fraction of all individuals in the population must apply mutually coherent classifications. These mutually coherent classifications are conventions i.e.,~reproduced over time due to the weight of precedent. Therefore, like conventions, norms can only be changed through collective action \citep{marwell1993critical}.

Individuals not directly involved in a particular interaction nevertheless classify it along normative dimensions and may become motivated to sanction as a result \citep{fehr2004third}. As such, norms are not sensitive to personal relationships between individuals or idiosyncratic beliefs or the knowledge of individuals. It is the individuals who represent, remember, and apply norms, but their force comes from the fact that many individuals represent, remember, and apply them in a similar way.

Norms depend critically on context. Since our basic logical reasoning skills are themselves normative behaviors, our theory predicts substantial context dependence in human basic logical reasoning skills. Indeed, this is what has been observed. In abstracted laboratory tasks with unfamiliar stimuli (like the Wason selection task), participants often fail to use \emph{modus tollens} whereas in argumentative context, they commonly make use of and understand arguments employing that very same logical rule \citep{mercier2011humans}.

\subsection{Implicit vs explicit norms}
\label{section:implicitVsExplicit}

A core part of our theory is the distinction between implicit and explicit norms. They differ in their representation, mechanism of action, and acquisition process. Explicit norms are articulable as verbalizable rules in standard language (e.g., laws, regulations, proverbs). They can be precisely formulated and used in logical or quasi-logical reasoning (e.g., legal adjudication). In contrast, implicit norms cannot be precisely articulated as a single standard in natural language. Instead, they are understood through examples and perceived appropriateness in context (e.g., conversational distance). Individuals sharing a culture generally agree on instances of pro-/anti-norm behavior guided by implicit norms, yet remain unable to articulate a consistent underlying rule, a phenomenon characterized as ``moral dumbfounding'' \citep{haidt2001emotional}. Unlike explicit norms, which must be actively attended to in order to influence behavior, individuals may not know when their actions have been influenced by implicit norms. They must rationalize after the fact their motivations to enforce and comply with implicit norms. If never asked, they may act in coherence with their implicit norms without ever producing explicit reasons for doing so, or may admit very flimsy justifications when unlikely to face pushback \citep{haidt2012righteous, mercier2017enigma}.

The two kinds of norms map onto distinct parts of the cognitive architecture. Explicit norms, when newly acquired, are stored as assemblies in memory ($M_t$), corresponding to the hippocampal system, which is critical for initial memory formation \citep{squire2015memory}. They influence behavior when they are retrieved into the global workspace ($z_t$) as explicit conditioning information for the pattern-completion network ($p$).
Implicit norms, conversely, are embedded within the parameters (weights) of $p$ itself, likely stored in the neocortex, potentially in the prefrontal cortex (PFC), which is implicated in abstract rule-following and context-dependent social behavior \citep{wallis2001single, beer2003regulatory}. 

Implicit norms are thus created by automatizing often-repeated cognitive patterns \citep{shiffrin1977controlled}. Over time, through repeated activation and rehearsal (potentially during sleep replay \citep{peyrache2009replay}), explicit norms can consolidate into implicit norms, becoming assimilated into the weights of the neocortical pattern-completion network \citep{mcclelland1995there}. This mechanism explains the automaticity of much appropriate behavior (e.g., adjusting speaking volume indoors vs.~outdoors).

\subsubsection{Serial processing}
\label{section:implicitLogic}

While one may naturally think explicit norms implement serial processing and implicit norms implement parallel processing, this would be misleading. Both kinds of norm can trace out serial steps of processing where a sequence of related representations arise in the global workspace. Furthermore, not all serial cognitive processing is effortful. There are sequential habits of thought that become automatized through repetition, practice, and expertise. If $x$-like items tend to be followed by $y$-like items during a sequential logical processing task that an individual frequently carries out, then the sequence will get consolidated into $p$ and become an implicit norm \citep{russin2025step, loo2025llms}.

This allows for the automatization of complex cognitive strategies, including logical reasoning. For instance, consider an individual with multiple possible behavioral scripts that start from $x$-like items---in some cases they are completed with $y$-like items, and in other cases they are completed with $z$-like items. We can think of these possibilities as akin to ``solution strategies''. We may further stipulate that a particular $y$ is the result of applying a deductive inference rule to a particular $x$. For instance, $x$ could be the set of premises \texttt{``If it's raining, then the ground is wet''} and \texttt{``It's raining''}, while $y$ could be the logical conclusion \texttt{``The ground is wet''} derived by applying \emph{modus ponens}. And so it becomes possible to implement logic using implicit norms.

Of course, it is unnecessary for pattern completion-based solution strategies to correspond to well-defined logical rules like \emph{modus ponens}. The strategy could be to complete $x$ with a repetition of its first element, or with the item written in the brightest print, or an inference like the one we described in Section~\ref{section:examplePatternCompletion}. Any such rule or pseudorule may become an implicit norm if repeated sufficiently often in practice.

Typical WEIRD (Western, Educated, Industrialized, Rich, and Democratic) adults in psychological studies have already been extensively exposed to formal logical rules in school curricula, and so have already consolidated implicit norms related to logic-based solution strategies. These strategies have been automatized \citep{shiffrin1977controlled}, making logical reasoning more intuitive \citep{deneys2019logic}. This does not mean that all ``textbook correct'' steps of reasoning are carried out one-by-one in every instance. Individuals may skip steps, substituting a hunch to get past a difficult step, or even work by just positing the final answer then subsequently seeking to discover the logical steps to reach it through \textit{post hoc rationalization} \citep{cushman2020rationalization}. Regardless of the particular strategy, and independently of how complicated it appears to be, it can always be automatized by sufficient practice and expertise, becoming a habit of thinking in a certain way.

\subsubsection{Precedence}
\label{section:precedence}

In cases of conflict, implicit norms appear to take precedence over explicit norms. This may be explained by noting that implicit norms are relatively recalcitrant to verbal reasoning \citep{haidt2001emotional}. When explicit step-by-step deliberation about normative appropriateness does occur, sometimes additional considerations activated by the context may neutralize a sense of obligation \citep{schwartz1977normative}. For instance, a person may initially feel obligated to donate money toward alleviating poverty in a far away place, but then after considering a rationale for why doing so is not their responsibility (e.g.~\texttt{``others who are much wealthier do not donate''}) conclude they have no such obligation. In the case of implicit norms, neutralization via rationalization is much less likely. You cannot reason your way out of a position maintained in an implicit norm. Implicit norms do change, but only over the slower timescale of consolidation. And they are no more likely to change in the direction of sound arguments than unsound. Repetition changes implicit norms, not reflection. In computational models of our theory, the reason we see these effects is the same reason that alignment methods that modify the weights of the LLM are a much more robust way to achieve a desired behavior than simply prompting the LLM to behave in the desired way \citep{ouyang2022training, kobis2025delegation}.

\subsubsection{Chains of thought: short and long}
\label{section:dualProcess}

Over the past half century, many dual process models of social cognition have been proposed. As a whole, they propose one system for fast, intuitive, and non-conscious judgments (implicit social cognition) and another for slow, deliberative, and conscious ones (explicit social cognition) \citep{greenwald1995implicit, wilson2000model, uleman1989unintended, kahneman2011thinking}. Although these are often considered to be separate systems, perhaps with their own separate neural substrates \citep{lieberman2007x}, alternative models have been proposed that are strongly aligned with the LLM processes discussed in this paper. Most notably, inspired by the Complimentary Memory Systems approach, \cite{cunningham2007attitudes} (see also \cite{cunningham2007iterative}) proposed that all forms of social cognition arise from the same neural network weights, but that the amount of time or reflection devoted to a task is what differed between "implicit" and "explicit" responses. Some judgments can be made relatively quickly and effortlessly, whereas other require retrieving additional context and considerations. Using this conceptualization, we can model `fast' processing effects (i.e.~``intuition'' or ``system 1'') with short autoregressive chains of thought or immediate responses. And, we can model `slow' processing effects (i.e. ``deliberation'' or ``system 2'') with long autoregressive chains of thought.

Our theory thus captures the difference between fast and slow processing by construction: we interpret ``deliberation'' as the generation of a long, explicit chain of thought that unfolds serially through time, with each step occupying the transient global workspace. A cognitive load task, such as rehearsing digits, functions as a persistent competing process. This secondary task must repeatedly re-activate its own content, recurrently co-opting the global workspace and thereby interrupting the serial generation of the primary deliberative chain of thought. This makes it difficult to sustain the fragile, step-by-step sequence required to apply an explicit norm. In contrast, implicit norms---which are consolidated into the weights of $p$---can generate an appropriate response using a much shorter autoregressive sequence. Implicit norms do not depend on the long, uninterrupted chains of thought that are disrupted under load or in speeded response. Therefore, cognitive load or speeded response paradigms are interpreted as isolating the output of the automatic, consolidated implicit norms. Consistent with this picture, \cite{gutierrez2007anger} showed that rationalization effects in moral judgment disappear under cognitive load conditions while preserving the judgment itself, suggesting judgments are based on implicit norms while their rationalization depends on in-context adaptation and retrieval of explicit norms from memory. In computational models, longer chain of thought sequences have been shown to sometimes degrade the performance of LLMs, analogously to overthinking in humans \citep{su2025between,wu2025more,chen2024not,shojaee2025illusion}.

It is instructive to consider how these models work in context of the Cognitive Reflection Test (CRT) \citep{frederick2005cognitive}.

\begin{equation*}
    (o_t, z_t) = \begin{bmatrix}
\texttt{A bat and a ball cost \$1.10 in total.
}\\
\texttt{The bat costs \$1.00 more than the ball.}\\
\texttt{Question: How much does the ball cost?}\\
\texttt{Answer:}
\end{bmatrix}.
\end{equation*}
One possible and ``intuitive'' (but incorrect) response would be given by the continuation $a_t \sim p (\cdot|o_t, z_t)$ could be
\begin{equation*}
a_t = \begin{bmatrix}
\texttt{10 cents}
\end{bmatrix}.
\end{equation*}

The way to get the less intuitive, but correct, answer would be through the longer ``chain of thought'' continuation. For example 
\begin{equation*}
a^\prime_t = \begin{bmatrix}
\texttt{OK, let's think about this step by step.}\\
\texttt{I need to solve two simultaneous equations:}\\
\texttt{bat = ball + \$1, and} \\
\texttt{bat + ball = \$1.10} \\
\texttt{OK, I see now, the answer is 5 cents}
\end{bmatrix}.
\end{equation*}

This CRT example can be mapped directly to the cases described above. The \texttt{10 cents} answer represents a fast, high-probability pattern, perhaps arising as a consequence of an implicit norm for simple subtraction problems. The correct \texttt{5 cents} answer is the product of a longer step-by-step reasoning process.

An individual's global workspace might simultaneously contain the \texttt{5 cents} candidate, the \texttt{10 cents} candidate, and the latter's chain-of-thought rationale. In this scenario, computing the pattern completion by sampling $p$ has the effect of arbitrating between the two candidates. It may favor the candidate attached to the justifying rationale, as in the argumentative norms of \cite{mercier2011humans}. This reconceptualization aligns with recent proposals to refine dual-process models \citep{deneys2023advancing, deneys2025defining}. A key finding in this literature is that individuals often intuitively detect conflict between an `intuitive' and a `correct' response, even when they ultimately provide the intuitive (incorrect) answer. In our theory, this `intuitive conflict detection' can be understood as the pattern-completion network $p$ assigning significant probability to multiple, mutually exclusive completions. For instance, in the CRT, the prompt may simultaneously activate the \texttt{``10 cents''} pattern and the \texttt{``bat = ball + \$1''} pattern. This competition results in a high-entropy probability distribution over the next step, which is registered cognitively as a sense of conflict or uncertainty. `Deliberation' (i.e., generating the long chain of thought) is thus not the action of a separate `system 2', but rather a specific, learned habit of thought that the actor employs to resolve this high-entropy state and converge on a single, defensible pattern.

Like other accounts, our theory predicts that simply engaging in cognitively demanding work is no guarantee that correct answers will be reached. In our theory, cognitive demand is simply the length of the autoregressive sequence, how long it goes before reaching a `stop symbol' (i.e.~a cell assembly that tells the brain to stop deliberating and output an answer). We turn next to discuss the situation of motivated reasoning where chains of thought become more and more biased the longer they grow.

\subsubsection{Motivated reasoning}
\label{section:motivatedReasoning}

Dual process theories have always been attractive partly due to the optimistic-sounding gloss they suggest: that we only need to think harder for our rational self to find the light of Truth and use it to correct our biased and careless base self. As a result of this background, dual-process theories generally have trouble accounting for effects where deliberation engenders more bias, not less \citep{mercier2011humans}.

By modeling identity-protective motivated cognition \citep{kahan2013ideology, kahan2017motivated} with our theory, we illustrate the working of a different mechanism: in this case the generation of a chain of thought is not blocked like in the cognitive load / fast response models, but instead it is steered or co-opted by a powerful implicit norm to produce motivated reasoning.

Expertise in a particular domain normally improves decision making in that domain. However, there are some situations where deliberation increases bias rather than decreasing it; e.g., political partisanship can make you bad at math when the math gives results that run counter to your prior politically selected belief~\citep{kahan2013ideology, kahan2017motivated}. In these cases, expertise can have the counterintuitive effect of producing worse decisions \citep{kahan2013ideology, kahan2017motivated}.

Consider how our theory models a participant in a motivated numeracy experiment \citep{kahan2017motivated}. In this paradigm, participants are shown data from a fictional study, framed either neutrally (e.g., about a skin cream's effectiveness) or in a politically charged manner (e.g., about a gun control measure's effect on crime). The data is presented in a $2\times2$ contingency table that is designed to be misleading; a superficial reading suggests one conclusion, while the correct mathematical calculation supports the opposite. A politically-motivated participant's chain of thought, when faced with results that challenge their identity, might be completed by their pattern-completion network as follows:
\begin{enumerate}
\item \texttt{The problem asks if a ban on guns decreased or increased crime in cities. My political identity is strongly tied to the belief that gun control is effective.}
\item \texttt{Looking at the table, more cities with the ban saw a decrease in crime than an increase. This fits my prior belief. This seems to be the answer.}
\item \texttt{Wait, to be sure, I should calculate the ratios, which is the proper way to evaluate this. Let's see... the ratio of cities where crime decreased is actually higher for the ``no ban'' group than the ``ban'' group.}
\item \texttt{Hmm, that's a counterintuitive result. My calculation must be wrong, or more likely, I'm misinterpreting what the numbers signify. The superficial reading from step 2 felt more straightforward and was consistent with the well-established fact that gun control works.}
\item \texttt{Ah, I see the likely flaw in my reasoning so far. The group of cities that didn't have the ban was likely different to begin with, so comparing simple ratios is not valid. The more robust rationale to use here is simply that a larger absolute number of cities with the ban saw improvement.}
\item \texttt{Therefore, the data shows the gun control ban was effective.}
\end{enumerate}
In this trace, the initial identity-based norm (Step 1) steers the entire reasoning process. When a conflict arises with a culturally-acquired explicit norm (i.e.,~\texttt{``I should calculate the ratios, which is the proper way to evaluate this''} in Step 3), the conflict is not resolved by abandoning the prior belief. Instead, the actor's $p$ generates a series of rationalizations (Steps 4 and 5) that dismiss the mathematical evidence. As found by \cite{kahan2017motivated}, greater numeracy (expertise) can make this rationalization easier, as it provides more sophisticated tools for Step 5.

Consistent with this account, \cite{taber2016illusion} showed that people spend more time and effort deliberating on arguments that contradict their prior beliefs and \cite{mercier2017enigma} argues that people rapidly and uncritically accept arguments that align with their prior beliefs. In our account, we treat such ``motivated reasoning'' phenomena as caused by explicit decision logic getting recruited by an implicit norm (connected to a social identity) to produce a sophisticated rationalization for an identity-protective decision. \cite{kunda1990case} suggests that people reason partly by searching through their memory specifically to find beliefs and explicit decision logics likely to support their desired conclusion when applied in the current context. Even the assimilation of new information may be affected by motivated cognition. Incoming sensory information may be incorporated into pre-existing and sometimes procrustean schemas before impacting any further cognition \citep{greenwald1980totalitarian}. Likewise, in our theory, activated identities, with their attendant implicit norms, dictate which assemblies enter the global workspace in subsequent serial processing and which explicit reasoning strategies get deployed.

\subsection{Norms as technologies}
\label{section:normsAsTechnologies}

Norms, which we regard as emerging from conventionalized patterns of sanctioning, can be understood as culturally evolved \emph{technologies} that address fundamental problems of coexistence and cooperation within a society \citep{ostrom1990governing, north1990institutions}. These normative technologies, ranging from property rights to rules of discourse, exhibit the characteristics of context-dependence, arbitrary origin, automaticity in application, and dynamic evolution, all of which are explicable through the lens of agent populations learning to reproduce patterns via their predictive networks ($p$). Their persistence and form are partially shaped by their functional efficacy within a given socio-ecological context, often emerging through undirected cultural evolution, potentially amplified by mechanisms like group selection \citep{wilson2013generalizing, henrich2004cultural} or deliberate design \citep{sunstein2019change}.

Crucially among these are \emph{epistemic norms}, which serve as cognitive technologies defining what a community regards as valid arguments, credible evidence, and reasonable hypotheses \citep{rorty1978philosophy, mills1940situated}. These norms are not universal, but culturally contingent, evolving through processes like norm commentary and paradigm shifts \citep{heyes2022rethinking, kuhn1962structure}. What a society designates as ``objective knowledge'' is therefore an output of its prevailing, socially constructed epistemic norms. The predictive pattern completion framework accounts for this by positing that an individual's $p$ internalizes these epistemic patterns from their social environment. The perceived ``usefulness'' or ``truth'' of certain epistemic norms can be reinforced if they lead to successful prediction or goal achievement, particularly in societies where epistemic norms evolved to rely on feedback from the ``non-human'' world (i.e.~the biophysical environment) for determinations of normative correctness, as in the institutions and norms of the scientific community.

This perspective allows us to understand rationality not as adherence to universal logical axioms or utility-maximization principles, but as behavior---including cognitive acts like reasoning and judgment---that aligns with the dominant epistemic norms of an individual's culture and context. Rationality, in this descriptive view, becomes a culturally-contingent virtue, an ideal to be aspired to, sometimes mirroring idealized public deliberative processes (cf.~\cite{hampshire1999justice}). Because the standards for such processes (e.g., what constitutes ``due weight'' or ``relevant evidence'') are themselves normative and culturally transmitted, this conception of rationality is inherently contingent, dynamic and context-dependent. An action or thought $y$ is deemed ``rational'' if it is a probable completion $y \sim p(\cdot|x)$ where the context $x$ includes relevant observations and the currently active epistemic norms, either explicitly in context or acting implicitly via the weights of $p$ due to prior consolidation. In this way, even that which is rational is ultimately controlled through sanctioning by members of the relevant community.

\subsection{Norm change}
\label{section:normChange}

Norms are collectively enacted, changing through both undirected processes like evolutionary drift \citep{young1993evolutionary} and strategic intervention \citep{finnemore1998international}. Appropriateness is not an inherently conservative force. Demands for reform are often driven by identity-based appeals, not just rational calculation \citep{march2011logic}. Change is frequently spurred by conflict between groups with differing normative preferences \citep{stastny2021normative, mukobi2023welfare}. The prevailing norm may shift due to evolving power dynamics \citep{young2015evolution, koster2020model}, strategic conflict resolution \citep{rahwan2018society, sunstein2019change, noblit2023normative}, or bandwagon effects where sanctioning behavior cascades \citep{vinitsky2023learning} (e.g.~rapid changes in the appropriateness of smoking cigarettes in public were mediated by informal sanctioning~\citep{alamar2006effect}). Such positive feedback can create tipping points: when enough individuals perceive the costs of challenging the status quo have fallen, rapid change can ensue \citep{crandall2018changing}. Consequently, while some norms exhibit substantial inertia \citep{bicchieri2016norms}, others, like mask-wearing during the COVID-19 pandemic, can emerge and spread with remarkable speed \citep{yang2022sociocultural}.

The mechanisms of change differ for explicit and implicit norms. Explicit norms, such as laws, are tied to institutions with established processes through which their rules can be changed \citep{hadfield2012law, hadfield2014microfoundations}---e.g.~when legislators perceive a need to change a law they can pass new legislation to do so. These systems possess what \cite{hart1961concept} called secondary rules---rules for making and altering other rules. Public ``norm commentary'' also propels the evolution of explicit norms by facilitating social learning and diffusing new normative interpretations \citep{heyes2022rethinking}.

Thomas Kuhn's account of scientific revolutions is an example of how epistemic norms are transformed by social mechanisms \citep{kuhn1962structure}. Kuhn argued that competing paradigms are not governed by a common logic. Rather, they operate using different vocabularies and incommensurable standards of evidence and reasoning \citep{rorty1978philosophy}. A paradigm shift is therefore not a belief update but a normative change affecting the way scientists reason. Like all norm changes, paradigm shifts are driven by collective action. Before the shift, a dominant paradigm is upheld by a conventional pattern of sanctions---peer review rejects papers that violate core assumptions, and funding is denied for unconventional research. A new paradigm eventually triumphs by winning a social struggle, e.g.~by establishing its own journals and communities of reviewers, creating a new pattern of social approval: a new set of normative practices for the newly formed scientific community.

This framework also accounts for the long-term, gradual evolution of implicit norms---the evolution of that which goes unspoken. This process shapes what can be understood as the ``history of common sense'' (e.g., \cite{rosenfeld2011common}). Our theory suggests a concrete computational mechanism for this. The predictive patterns consolidated in $p$ are a direct reflection of the historical and cultural context in which they were ``trained.'' For example, an LLM trained exclusively on a corpus of text and social data from 1950 would inevitably consolidate a vastly different set of implicit norms regarding gender relations, authority, and public discourse than one trained on data from 2025. In this way, slow, inter-generational cultural change is instantiated as a fundamental shift in the underlying predictive patterns that constitute common sense itself.

\section{Explaining the stylized facts}
\label{section:explainingStylizedFacts}

We are now able to spell out our theory's proposed explanation for each of the stylized facts of appropriateness we articulated as our explanatory targets in Section~\ref{section:desiderata}.

\begin{enumerate}
    \item \textbf{Context dependence}---everything noticed about the current situation is in the the global workspace and conditions the choice of how to behave. Additional information about the identity of the actor is also always in the global workspace, and social identity has a large patterning effect on appropriate behavior across a large number of situations and roles. Moreover, each culture may consider a different set of behaviors, dress, and demeanor to be appropriate for each situation.

    \item \textbf{Arbitrariness}---response features of no material consequence may be moralized. There is no requirement for conventions and norms to be fair or efficient. While some norms are likely non-arbitrary, such as those critical for facilitating cooperation, the mechanisms by which norms emerge and are maintained pose little constraint on their content. The arbitrariness of norms is inherited from the arbitrariness of conventions. Because conventions are arbitrary in the sense of Section~\ref{section:conventionsDefinition} it is thus possible for arbitrary patterns of sanctioning to become conventional, i.e.~to generate norms. 

    \item \textbf{Automaticity}---In our model, the cognitive effort required for a decision corresponds to the length of the ``chain of thought''---the sequence of pattern completions needed to arrive at an action (Section~\ref{section:dualProcess}). Most appropriate behavior is \textit{automatic} because it is the result of a short and highly probable pattern completion, prompted by the current context. For instance, speaking quietly indoors does not require deliberation; it is a direct, high-probability completion. In contrast, \textit{deliberative} or effortful thought occurs when a longer chain of reasoning is required to resolve ambiguity or conflict between competing patterns. This distinction explains why acting appropriately is usually, but not always, automatic.
         
    \item \textbf{Dynamism}---Societal change is driven by positive feedback dynamics as everyone shifts to play their part of a newly emerging status quo. We discussed several mechanisms including both deliberate collective action and more emergent bottom-up processes like the bandwagon effect in Section~\ref{section:normChange}. Norm change can  proceed quite rapidly with strong positive feedback that could unleash cascade dynamics.

    \item \textbf{Sanctioning}---This was a direct consequence of how we defined norm in Section~\ref{section:normsDefinition}. We define a behavior as normative, and therefore appropriate with strangers, precisely when it is encouraged or discouraged by a conventional pattern of social sanctioning. Sanctioning is not merely a feature of norms---it is the very engine that generates them. Sanctions give norms their binding force, separating them from mere conventions.
\end{enumerate}

\section{Pattern completion is all you need: Rationality as a normative practice}
\label{section:patternCompletionIsAllYouNeed}

Throughout this paper, we have framed our theory as an alternative to rational-choice theories that posit maximization of a one-dimensional reward or utility as the fundamental driver of behavior. We now complete the argument. We show that, in our theory, the rational logic of making choices by considering their expected consequences may be regarded as a specific, culturally-contingent, \emph{normative practice}---just one specific decision logic among many possibilities.

This re-framing reflects a foundational difference in modeling philosophy. Rational-choice theories depend on \textbf{exogenous} inputs: the modeler must specify tastes, preferences, or a utility function from outside the model before it can run. These externally-defined preferences are treated as the primordial causes of all subsequent behavior. Our theory, in contrast, treats preferences and goals as \textbf{endogenous} phenomena that emerge from the interaction of the model and its social environment. Thus, preferences are not given and fixed but instead constructed from the internal dynamics of the model, reflecting the cognitive processing of an encultured agent.

We propose that the logic of rational choice is a culturally evolved cognitive technology, an epistemic norm that establishes a community's standards for good decision-making. The procedure we outlined as Decision Logic A (Section~\ref{section:decisionLogics})---identifying options, evaluating consequences, and selecting the one with the highest expected value---is an explicit norm: an articulable script for decision-making institutionalized in specific communities, such as economics and engineering. An individual ``acts rationally'' just as they may be guided by any other explicit norm. The rational-choice procedure is acquired through instruction and maintained by conventional patterns of sanctioning. An economist who submits a paper that violates the axioms of expected utility theory will be sanctioned and their paper will be rejected. A manager who makes a decision without a defensible cost-benefit analysis may face professional sanction. In these contexts, applying the logic of rationality is the appropriate thing to do. The behavior is normative because it is sustained by a conventional pattern of sanctioning within a community. Like all explicit norms it may also be consolidated through repeated use, thereafter functioning like an implicit norm, automatically.

This is not to say that complex, goal-directed expertise, such as that invested in becoming a master craftsman, is not a form of rationality. However, our theory suggests that the underlying cognitive process for such expertise may be fundamentally different from the one articulated in Decision Logic A. While an external observer can \emph{model} the craftsman's behavior `as-if' they are solving an optimization problem (e.g., ``minimizing steps''), the craftsman's internal process is more likely an application of Decision Logic B: ``What does a master \emph{such as I} do in a situation \emph{such as this}?'' (or its consolidated implicit norm equivalent). This expertise is often better understood as the application of a vast, culturally-transmitted body of knowledge, which is often arbitrary and non-obvious \citep{boyd2011cultural}. For example, traditional Hadza bow makers acquire their skill by following a complex, culturally-specific procedure, not by rationally inferring the principles of physics from first principles \citep{harris2021role}. Notably, expert Hadza bow makers systematically lacked knowledge of some trade-offs involved in designing a bow, and most study participants justified their design choices with comments like ``it is the Hadza way'' or ``this is the way the elders have instructed us''~\citep{harris2021role}---results which are just what we would expect for experts who work by applying a body of cultural knowledge to their craft, but would be mysterious if they were rational agents, approaching their craft via accurate world modeling. Their expertise, which in this view appears to be much more like Decision Logic B, is highly effective (they use the bows for hunting, and in this way provide most of the meat in their diet), but to mistake it for the universal application of Decision Logic A may be to view complex cultural knowledge through the limited lens of one's own, culturally-specific epistemic norms.

Crucially, our theory can readily produce ``rational'' behavior without requiring a built-in utility maximizer. This becomes clear when you consider how pattern completion operates on a context that includes an explicit goal. If an agent's global workspace contains the prompt, \texttt{``The goal is to maximize quarterly profits. The most effective action to take is:''}, its pattern-completion network $p$, having been trained on a corpus of business and economic texts, will complete the pattern with actions corresponding to profit-maximization strategies. The goal does not function as an external utility signal to be optimized, but as a symbolic constraint within the context that shapes the probability distribution over possible actions. The agent is not maximizing a reward function; it is parroting and completing a normative pattern associated with goal achievement. This demonstrates that goal-directed behavior is a straightforward consequence of the pattern completion mechanism, not a separate and fundamentally different process.

With this in hand, we can make our parsimony argument explicit. If our theory can reproduce all the behavioral phenomena explained by reward/preference-based theories---which it can, by treating rationality as a learnable normative script---then it offers a simpler explanation. Rational-choice theories require two sets of assumptions: the existence of a fundamental utility-maximization mechanism, and the exogenous specification of a utility function for every context. Our theory uses only one mechanism: predictive pattern completion. Because it explains the same data and, as we have argued, a much wider range of social phenomena with fewer foundational assumptions, then by Occam's razor, it ought to be the preferred theory.

In particular, our theory provides a more robust account of endogenous preference formation \citep{bowles1998endogenous}. Human preferences are not fixed but are dynamically shaped by experience, social interaction, and self-inference \citep{bem1967self, mackie1977ethics, stryker2007identity, cushman2020rationalization}. In our theory, an individual's ``preferences'' are constructed when their pattern-completion network $p$ generates inferences based on memories, social context, and identity. When an actor's global workspace contains the assembly \texttt{``I am the kind of person who likes rock music''}, this is not a report on a pre-existing utility function; it is the effect of a pattern completion over past experiences, shaped by the social norms of the groups to which they belong. Preferences are not pre-social primitives that cause behavior; they are the consequences of an ongoing process of socially-informed, cognitive self-constitution \citep{bourdieu1984distinction}.  Notice that the path-dependence of preference formation is clear: different early experiences may set the individual on an entirely different, yet equally consistent, preference trajectory.

Our theory provides a more natural account for a great many phenomena which are challenging for standard reward/utility-based theories, such as the influence of norms controlling behaviors that would otherwise have no material consequence (e.g.~food taboos for harmless foods \citep{navarrete2003meat, mcelreath2003shared}), the influence of mere familiarity on preference \citep{zajonc1968attitudinal}, and the observation that people uphold norms because they participate in a particular way of life, rather than the reverse \citep{mackie1977ethics}.

This reconceptualization of rationality as a norm is an expression of the `society-first' approach we articulated in the introduction to this article. Most cognitive and economic models are implicitly `individual-first': they begin with an agent endowed with a fixed, pre-specified cognitive architecture---such as reward maximization---and then seek to explain emergent social phenomena from the interactions of many such agents. In these models, the agent's fundamental rationality is an axiom, not a phenomenon to be explained. Our theory reverses this explanatory priority by making rationality an effect of social order, not its cause. Our theory differs from others in that it explains the crucial link that is often assumed but usually left unexamined: the process by which macro-level cultural constructs, like the explicit rules for logical reasoning become internalized to shape the micro-level cognitive processes of an individual. In our theory, the social environment provides the training data and normative context that trains the individual's pattern-completion network. Therefore, our theory does not just take the rational agent as a given; it explains how a person learns to become one by internalizing the epistemic norms of their community. This provides a more complete theory of cognition by accounting for the cultural constitution of the agent itself.

This `society-first' theory also contrasts with `revealed preferences' and `inverse RL' approaches, which do not require specifying utilities \emph{a priori} but instead infer their existence from an agent's expressed behavior \citep{ng2000algorithms}. While this approach is powerful, it relies on a critical assumption of a stationary reward or utility function. This makes it difficult to account for endogenous preference change. For instance, if an individual's behavior shows a preference for rock music at one time and electronic music at a later time, the `revealed preferences' model would fail, as it violates stationarity. The straightforward explanation---that the person simply \emph{changed their mind} or manifested a new preference in a new context---is incompatible with the model. Our theory, in contrast, handles this naturally: treating `preferences' as the \emph{effect} of a pattern completion process over a dynamic set of memories and the current social context. As the context and memories change, the `preference' (i.e., the most probable pattern completion) also changes.

\section{Discussion}
\label{section:discussion}

In this paper, we proposed a society-first theory of social order in which individuals are modeled as culturally pretrained actors whose decision-making relies on predictive pattern completion, an operation analogous to that of an LLM. We designed this theory to explain how social order emerges from individuals learning and adhering to local, context-dependent norms. The theory posits that individuals generate behavior by completing symbolic sequences in a global workspace, effectively answering the question, ``What does a person such as I do in a situation such as this?''. We argued that this mechanism provides a unified explanation for five stylized facts of human normative behavior: its context-dependence, arbitrariness, automaticity, dynamism, and enforcement via social sanctioning.

The theory distinguishes between \textit{explicit norms}, articulable rules processed in-context, and \textit{implicit norms}, automatic behaviors consolidated into the cognitive architecture, thereby accounting for why behaving appropriately is normally habitual, and only rarely relies on deliberation. By defining conventions, sanctions, and norms in terms of predictive pattern completion, and eschewing the need to posit scalar reward signals or ordinal preference relations, our theory challenges rational-choice theories. If we can explain all the same phenomena without recourse to a brittle, exogenous scalar reward signal, our theory offers both greater parsimony and programmability. The prospect of providing simpler explanations for endogenous preference formation effects is especially important in this respect. This perspective culminates in a reconceptualization of rationality itself, not as universal utility maximization, but as adherence to specific culturally evolved norms.

Unlike traditional cognitive science models, which usually model individuals in unrealistically undersocialized ways, we have sought to explain society as a collection of culturally competent actors. In our theory, individuals are initialized with cultural knowledge, roles, and expectations. This society-first approach to simulation is made possible through the use of new generative agent-based modeling platforms like Concordia \citep{vezhnevets2023generative} and similar systems (e.g.~\cite{park2023generative}). Each simulation instance is not a reinvention of society from scratch. Instead, simulations constructed in accordance with this theory are concerned with working out the consequences of socially patterned interactions, enabling researchers to test counterfactual histories \citep{underwood2025can}, explore alternative cultural configurations, and identify the minimal mechanisms necessary for complex phenomena to emerge \citep{bednar2007culture, miller2007complex}. Society, construed this way, also impacts the individual, generating their preferences, beliefs, and common sense. Thus society and individual are co-constituted: individuals draw upon social knowledge to act, and in so doing collectively reproduce and change society, operationalizing sociology's core meta-theoretical framework and linking it with cognitive science.\\



\bibliographystyle{abbrvnat}
\nobibliography*
\bibliography{refs}

\end{document}